# Assessment of Vehicular Vision Obstruction Due to Driver-Side B-Pillar and Remediation with Blind Spot Eliminator

Dilara Baysal

dbaysal@purdue.edu

## Abstract

Blind spots created by the driver-side B-pillar impair the ability of the driver to assess their surroundings accurately, significantly contributing to the frequency and severity of vehicular accidents. Vehicle manufacturers are unable to readily eliminate the B-pillar due to regulatory guidelines intended to protect vehicular occupants in the event of side collisions and rollover incidents. Furthermore, assistance implements utilized to counteract the adverse effects of blind spots remain ineffective due to technological limitations and optical impediments.

This paper introduces mechanisms to quantify the obstruction caused by the B-pillar when the head of the driver is facing forward and turning 90 degrees, typical of an over-the-shoulder blind spot check. It uses the metrics developed to demonstrate the relationship between B-pillar width and the obstruction angle. The paper then creates a methodology to determine the movement required of the driver to eliminate blind spots. Ultimately, this paper proposes a solution, the Blind Spot Eliminator, and demonstrates that it successfully decreases both the obstruction angle and, consequently, the required driver movement. A prototype of the Blind Spot Eliminator is also constructed and experimented with using a mannequin to model human vision in a typical passenger vehicle. The results of this experiment illustrated a substantial improvement in viewing ability, as predicted by earlier calculations. Therefore, this paper concludes that the proposed Blind Spot Eliminator has excellent potential to improve driver safety and reduce vehicular accidents.

## 1. Introduction

B-pillars are commonly associated with blind spots, which can cause vehicular crashes. The research in this paper will determine and propose practical ways to reduce and even eliminate the blind spot caused by the B-pillar located on the driver's left side and thus improve vehicle safety.

## 1.1. What is a Vehicular B-Pillar?

A vehicular B-pillar ("B-pillar") is a central supporting structure on both sides of the cabin, typically located near the front seats of a vehicle. This structure resides between a vehicle's front and back doors [1]. The B-pillar is generally composed of steel and is welded to a vehicle's roof panel, floor pan, and rocker panel [2]. Although there has been experimentation and development in recent years regarding the use of composite materials in B-pillar construction, further developments still need to be implemented for such materials to be successfully integrated into the consumer market [3]. Furthermore, vehicles that do not utilize B-pillars are subject to negatively affected structural integrity and related safety issues [4].

## 1.2. B-Pillars and Vehicular Blind Spots

B-pillars block the driver's field of vision and result in blind spots, portions of a vehicle's surrounding area that the driver cannot view. Vehicle manufacturers often utilize increasingly prominent and thicker B-pillars to improve the safety of vehicles in the event of a collision. Unfortunately, this remedy creates even larger driver blind spots [5]. To combat this adverse effect, various motor vehicle agencies, such as the California DMV, instruct drivers to monitor their blind spots by looking over their shoulders during traffic maneuvers that include but are not limited to changing lanes, merging or exiting traffic, and turning [6].

## 1.3. Blind Spots and Roadside Vehicular Crashes

As the number of people on the roads increases yearly, roadside crashes have correspondingly increased, highlighting the need for testing and research on crash causes and avoidance mechanisms [7]. It has been determined that most roadway crashes are due to blind spots and related visual problems that occur while driving [5]. Hazardous conditions ensue when the driver cannot observe vehicles, bikes, or pedestrians within such blind spots. In turn, drivers' ability to make appropriate real-time choices may be hindered, and vehicular crashes may result. B-pillar-induced blind spots positively correlate with a higher frequency in lane-change crashes [8].

## 1.4. The Importance of B-Pillars

While B-pillars contribute to driver vision obstructions, they also provide essential vehicular safety features. B-pillars are a critical barrier of protection against side crashes, constituting 23 percent of all vehicle casualties nationwide [9]. Furthermore, B-pillars are vital to vehicular roof support [4]. In particular, the National Highway Traffic Safety Administration of the United States has imposed regulations requiring roofs to support from 1.5 to 3 times the vehicle's weight [10]. B-pillar roof support is also crucial in the event of a vehicular rollover, which comprises 28 percent of all crash casualties [9].

In addition, B-pillars contribute to improved vehicular Insurance Institute for Highway Safety (IIHS) ratings [11]. This is particularly relevant concerning vehicular crashes due to lane changes, classified as side crashes. Vehicles with IIHS-enhanced side safety ratings tend to pose a reduced fatality risk for occupants. In left-side crashes, for example, driver mortality is reduced by approximately 70 percent of vehicles that earn an IIHS side safety rating of 'good' compared to 'poor' [12].

## 1.5. B-Pillar-Induced Blind Spot Reduction Efforts

Multiple technological advancements are being researched and implemented to reduce blind spots caused by B-pillars. Companies such as Volvo, Acura, and Mazda offer optional blind spot detection additions to their vehicles. These technologies are often lights on the side mirror that turn on when objects are detected within the driver's blind spot. Studies have shown that implementing side-view assist systems may prevent approximately 395,000 crashes annually [13].

Unfortunately, such technology has largely failed to significantly decrease crash occurrence rates directly related to B-pillar-induced blind spots. In vehicles equipped with blind spot monitoring systems, lane change crashes tended to decrease by only 14 percent with a 95 percent confidence limit [14]. Related technological limitations may contribute to this issue. For example, Volvo's website indicates that sensor obstructions hinder the efficacy of its Blind Spot Information System due to factors like dirt, ice, and snow. Furthermore, Volvo noted that its Blind Spot Information System might disengage when a trailer connects to a vehicle's electrical system [15]. In addition, technology that allows drivers to see behind pillars in real-time by utilizing eye tracking and an advanced camera system is currently in development for A-pillars. To date, no comparable technology is available for B-pillars [16].

B-pillar positioning can be an important factor concerning blind spot reduction. In four-door vehicle models, B-pillars tend to be positioned further forward than their two-door receding B-pillar counterparts. Indeed, two-door models have an increased visibility angle from the B-pillar of 28 degrees compared to four-door models [8]. Due to such pillar positioning, four-door models are more susceptible to prominent blind spots. In turn, four-door vehicles generally have a 17% higher chance of involvement in an accident than two-door vehicles [8].

The B-pillar composition can also influence blind spot incidence. Ongoing research is being conducted to determine how to produce B-pillars from composite materials, such as carbon fiber, that adhere to National Highway Traffic Safety Administration regulations [3]. Carbon fiber is approximately five times stronger and lighter than steel [17]. Although steel tends to be the primary component of contemporary B-pillars, utilizing carbon fiber instead can empower vehicle manufacturers to achieve a slimmer design that reduces pillar obstruction. However, carbon fiber is considerably more expensive per pound than steel [18]. In addition, the labor required to produce B-pillar components from carbon fiber is typically time-intensive and requires skilled human labor and costly specialized machinery [19]. These factors significantly increase carbon fiber production costs relative to steel and greatly reduce the appeal of carbon fiber B-pillars for manufacturers.

## 1.6. Blind Spot Mirrors Can Be Ineffective

One example of a commonly utilized blind spot detection system is the blind spot mirror. This is a convex mirror attached to a vehicle's side mirrors. The convex shape of such a mirror allows drivers to view their surroundings with a greater field of view, thus allowing them to see objects and obstacles that would ordinarily be occluded due to blind spots. However, there are many downsides to such a device. For example, convex mirrors can be optically misinformative. The convex shape of such mirrors increases the field of view but distorts the images, causing the objects reflected in these mirrors to seem smaller. [20] This can result in a misleading perception of drivers' surroundings in the form of overestimation, a belief that objects are further than they are in actuality. Overestimation has been attributed to vehicular crashes and other road accidents. [21]

In addition, reflected images seen through mirrors are horizontally flipped. This skewed orientation is contrary to the objects' natural alignment. Since objects appear in the mirror differently from how they would normally be perceived, the driver may pay less credence to them [22]. In turn, this could lead to an array of negative vehicular navigation issues – including crashes.

### 1.7. Visual Multitasking

Blind spot mirrors and other blind spot technologies require excessive visual multitasking of drivers. Unfortunately, this can diminish driver focus on the array of other tasks required to navigate a vehicle safely. When drivers check their blind spot mirrors, they split their visual focus and automatically add an additional task to their visual workload. In these situations, a driver's primary task becomes processing visuals from the blind spot mirror. Meanwhile, the driver's background task entails looking at the road in front of the car. This misprioritization may cause a delay in the execution of both tasks. Furthermore, it can significantly reduce performance when compared to drivers who dedicate their focus to one visual [23]. The impact on the visual workload is even more drastic when the background task is non-trivial, as would be expected when driving a motor vehicle in more challenging roadway conditions [24]. This increases risk as drivers may not be able to comprehend visual stimuli quickly. Unfortunately, inadequate assessment of blind spot clearance and reduced reactivity to driving situations in the direction of travel can result.

### 1.8. Focusing

Human eyes utilize binocular vision, in which muscles keep the right and left eyes' lines of sight fixed on the same point. The brain combines the images from both the left and right eyes [25]. Varied fixation of the eyes is required to view objects at different distances when sight lines from both eyes intersect. In a typical blind-spot check, the driver's eyes tend to transition from focusing on the road ahead via the front windshield to looking out the passenger side window while maintaining the same or similar fixation or focus. Indeed, both scenarios require a view of similar angles and distances. A blind spot detector, or blind spot mirror, is located on a vehicle's side mirrors. Furthermore, this tool requires drivers to adjust their focus to prioritize the detector or mirror, then quickly adjust back to the road ahead. This brief and intensive exercise that checks for vehicles in blind spots may prove difficult, especially for elderly drivers with presbyopia who require longer time spans to readjust their focus [26].

## 2. Methods

This paper derives equations to determine the obstruction angle due to the B-pillar when the driver's head is facing forward and when it is turned 90 degrees. This angle was utilized to simulate the driver's over-the-shoulder blind spot check. These equations are then used to determine the requisite distance that the driver's head would need to move to eliminate the blind spot. Another set of equations was derived by analyzing the visual impact of utilizing a refractive sheet placed on the driver's side window to the right side of the B-pillar. Results were tabulated and graphed to study the effect of B-pillar width on the obstruction angle and movement required to eliminate the blind spots. This demonstrated the effectiveness of the introduced refractive sheet.

## 2.1 A Metric for Driver's Side Blind Spots

This paper will demonstrate the extent of blind spots created by the B-pillar when drivers attempt to check for traffic and obstructions on the left side of their vehicles. Human anatomy causes vision and spatial cognition to be angular, with sight lines meeting at a focal point behind drivers' heads. Sight lines are indicators of an eye's vision, as they are drawn from the eye to the focus of the eye [27]. Thus, it is appropriate to quantify and label the blind spot created by an obstruction as an obstruction angle. First, an equation for the obstruction angle will be derived from instances wherein drivers check their blind spots without turning their heads. Then, an obstruction angle equation for when drivers turn their heads but remain stationary will also be derived. These equations will subsequently determine the extent of driver head movement required to reduce or eliminate blind spots.

Human eyes have a nominal viewing angle of 120 degrees. Binocular vision, where there is an overlap of both fields of view for each eye, is also approximately 120 degrees. The monocular vision for each eye, where there is no assistance from the other eye, is approximately 40 degrees [28].

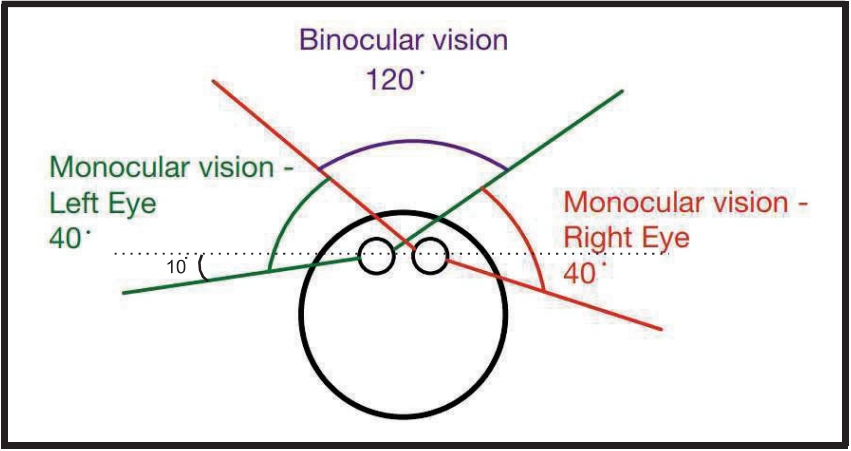

**Figure 1.** Areas of human vision.

Unless otherwise noted, this paper will refer to the default position of a driver as the driver sitting evenly in the middle of the driver's seat, facing straight ahead towards the direction of travel, head resting on the middle point of the headrest. The distance of the centerline of the headrest to the centerline of the B-pillar is defined as $h$. The horizontal distance from the left edge of the B-pillar to the middle point of the headrest, thus the middle of the back of the driver's head in the default position, is defined as $H_0$. This is illustrated in **Figure 2** below.

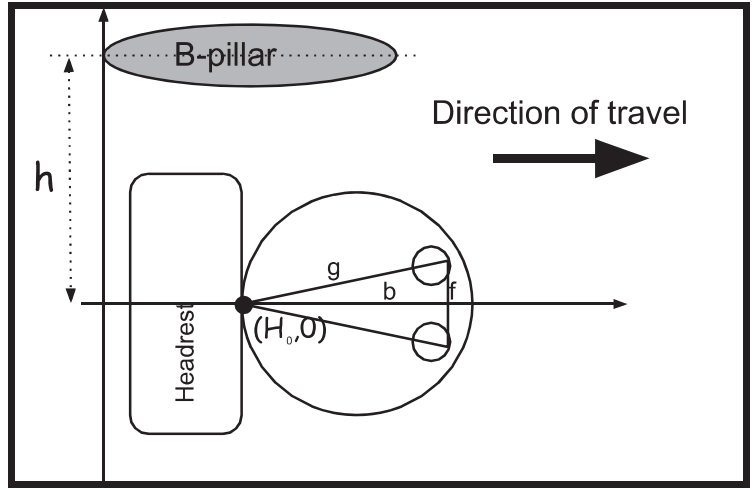

**Figure 2.** Driver's default position and orientation.

    The two pupils and the middle point on the back of the head form an isosceles triangle with two long sides of length $g$ and the distance between the two pupils, $f$. The latter is also called the pupillary distance and typically measures between 5.4 cm to 7.4 cm in adults [29]. The x-axis position of the pupils in the default position, $b$, can be found using the Pythagorean theorem.

$$b = \sqrt{g^2 - \frac{f^2}{4}}$$

    Drivers possess an additional visual aid: side mirrors. Drivers adjust their mirrors according to their preferences. Often, a driver of average height positions these mirrors at an angle such that the sight line closest to the vehicle brushes along the side of the car [30]. However, mirrors angled in this way only allow drivers to view a narrow slice of their surroundings relatively close to their vehicles.

    When looking at a side mirror to inspect the side and behind of the vehicle, drivers would rapidly move their eyes to the mirror's edges to maximize their visibility area. This is achieved when the right eye's sight line is reflected at the left edge of the mirror and vice versa for the left eye. In this way, the driver can maximize the reflection angle. This is illustrated in **Figures 3** and **4** below, with the driver obtaining a broader coverage with the right eye, yielding a greater reflection angle, κ, with the right eye.

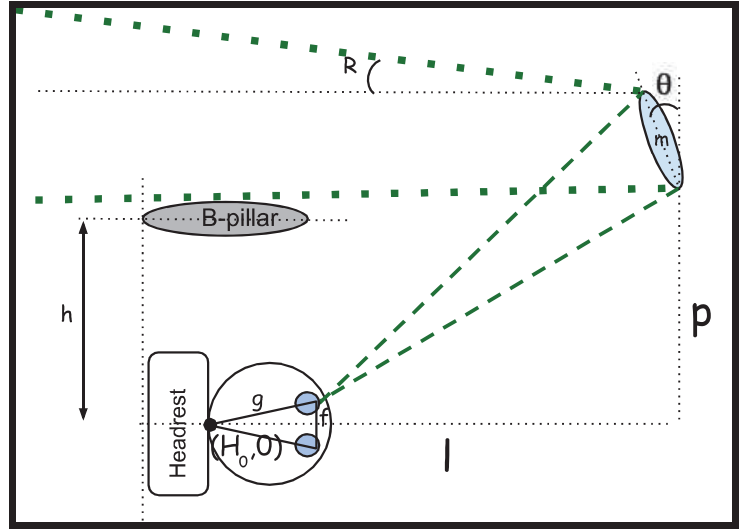

**Figure 3.** Mirror glance with left eye.

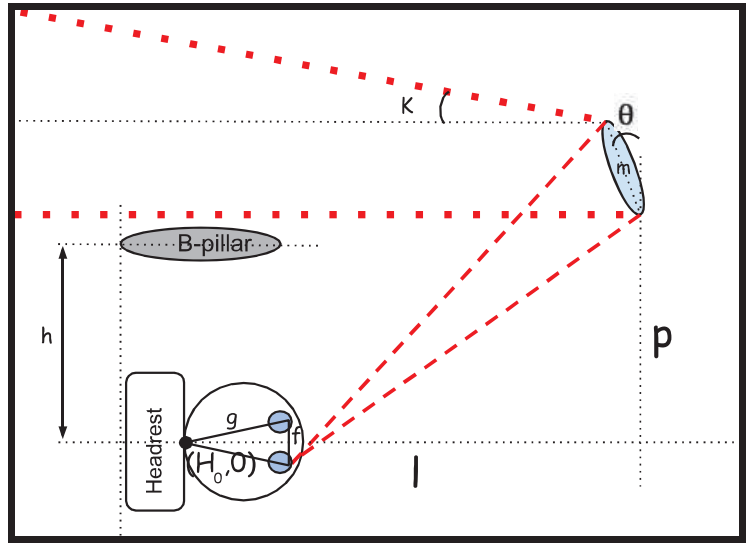

**Figure 4.** Mirror glance with right eye.

The first step in this process entails determining the angle of the outer sight line, κ. The angle of the mirror, θ, determines the value of κ and the side mirror's coverage area. Utilizing the law of reflection, which states that light will reflect off a mirror at the same angle that it arrives at the mirror (commonly known as incidence), the following diagram is derived [31].

The distance of the base of the side mirror to the centerline of the headrest is defined as $p$ and, for many vehicles, this value is slightly larger than $h$, the distance of the B-pillar centerline to the center of the headrest. The distance of the mirror's base to the left edge of B-pillar is referred to as $l$. Thus, keeping the origin at the center of the headrest, the base of the side mirror is at the position $(l, p)$. The length of the side mirror is defined as $m$.

The following diagram that shows the reflection of the right eye sight line can be devised to determine the angle of κ, the outer sight line's angle with the horizontal axis, pictured below.

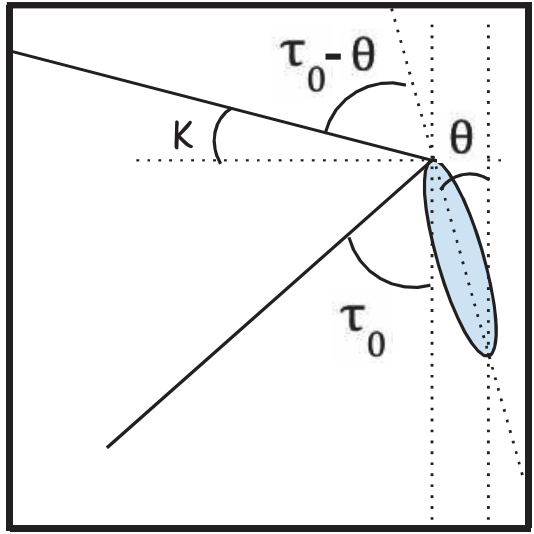

**Figure 5.** Close-up of sight line reflection at the side mirror.

The angle of the mirror's top sight line is κ and θ is the angle of the mirror. $\tau_0$ can be determined from the right triangle the right eye makes with the mirror.

$$\tau_0 = arctan\left(\frac{l - H_0 - b - m \cdot sin(\theta)}{\frac{f}{2} + p + m \cdot cos(\theta)}\right)$$

$$\kappa = 90 - \tau_0 - 2\theta$$

Once κ is known, a slope equation for the mirror sight line can be developed.

$$y_m = -tan(\kappa) \cdot (x_m - (l - m \cdot sin(\theta))) + m \cdot cos(\theta) + p$$
$$y_m = cot(2\theta + \tau_0)(l - m \cdot sin(\theta) - x_m) + m \cdot cos(\theta) + p \qquad (1)$$

## 3. B-Pillar Blind Spot Head Stationary at Different Positions - Monocular

Without a head turn, the driver's left side is viewed by the monocular vision of the driver's left eye. The B pillar blocks a significant part of the view, as seen in the diagram below. Sight lines for the left eye consist of the left and right sight lines. The left sight line is shown below as ending within the B-pillar at the maximum angle that the 120-degree range of vision will allow, and the right sight line is shown as connecting the right side of the B-pillar to the eye. These two lines demonstrate the maximum range that the left eye can see.

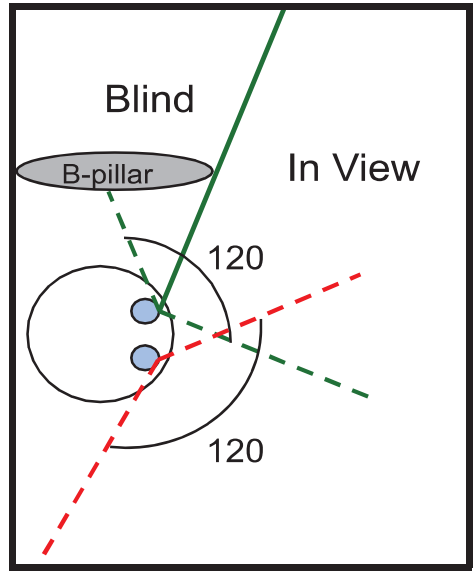

**Figure 6.** B-pillar blocking view in the default position.

The obstruction angle is the angle of the area beyond the bounds of the driver's sight lines. Even with the mirror use, the obstruction angle remains the same, as the sight line is unchanged past the bounds of the mirror view. λ is defined as the obstruction angle for monocular viewing and ρ is the angle of vision to the horizontal axis.

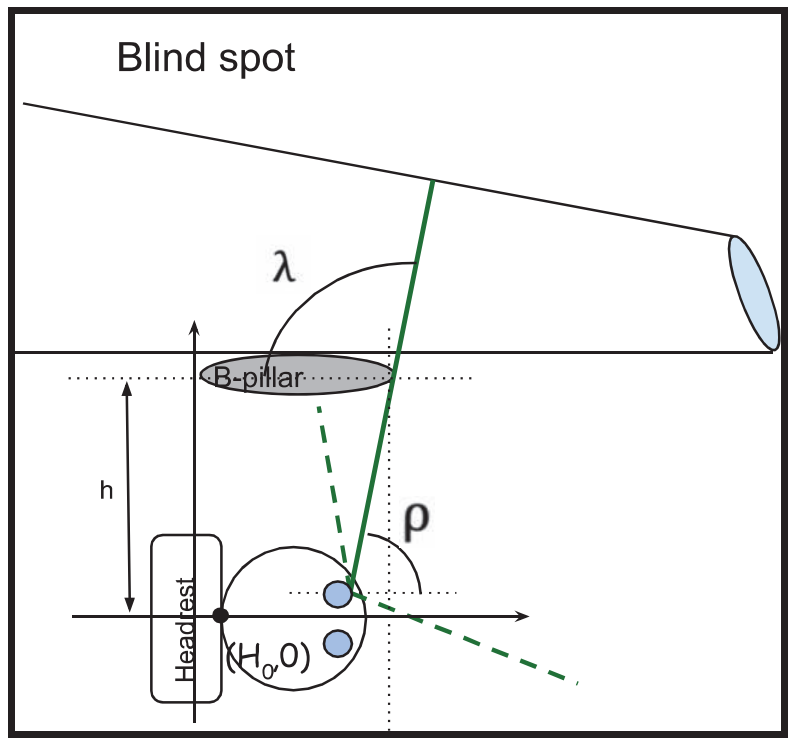

**Figure 7.** Obstruction angle with monocular vision.

An equation to calculate λ can be derived as follows.

$$\rho = arctan(\frac{h-\frac{f}{2}}{W-H_0-b}) = 180^o - \rho$$

$$\lambda = 180^o - arctan(\frac{h-\frac{f}{2}}{W-H_0-b}) = arctan(\frac{h-\frac{f}{2}}{H_0+b-W}) \ (mod \ \pi) \qquad (2)$$

## 4. B-Pillar Blind Spot Head Stationary at Different Positions - Binocular

As previously mentioned, particularly in North America, new drivers are instructed to reduce or eliminate the vision gap due to the aforementioned blind spot by rotating their heads and glancing over their shoulders. This study will determine how rotating the head 90 degrees, while keeping the head stationary, affects the resulting blind spot angle experienced by the driver due to the driver-side B-pillar.

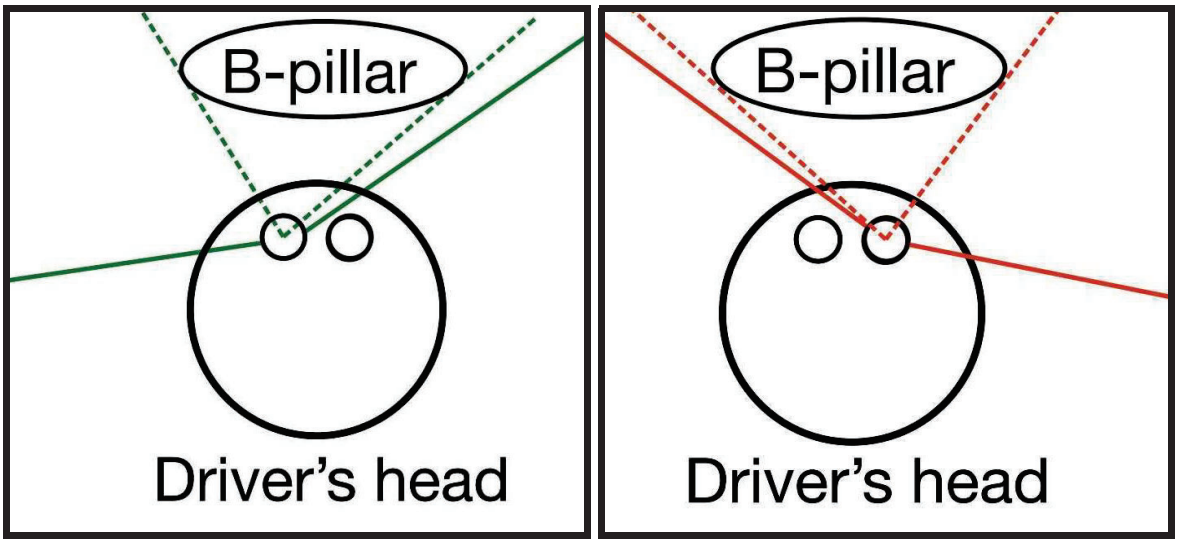

**Figures 8, 9.** Binocular vision during over-the-shoulder view.

Within the 120-degree angle for the individual eyes, lines called sight lines were drawn to indicate where each eye was obstructed by the pillar [27], and superimposed over each other, shown below as dotted lines. Because eyes compensate for each other to create a single percept during a process called binocular fusion [25], only the overlap of the individual eyes' obstruction was considered for the total angle of obstruction, α, as shown below.

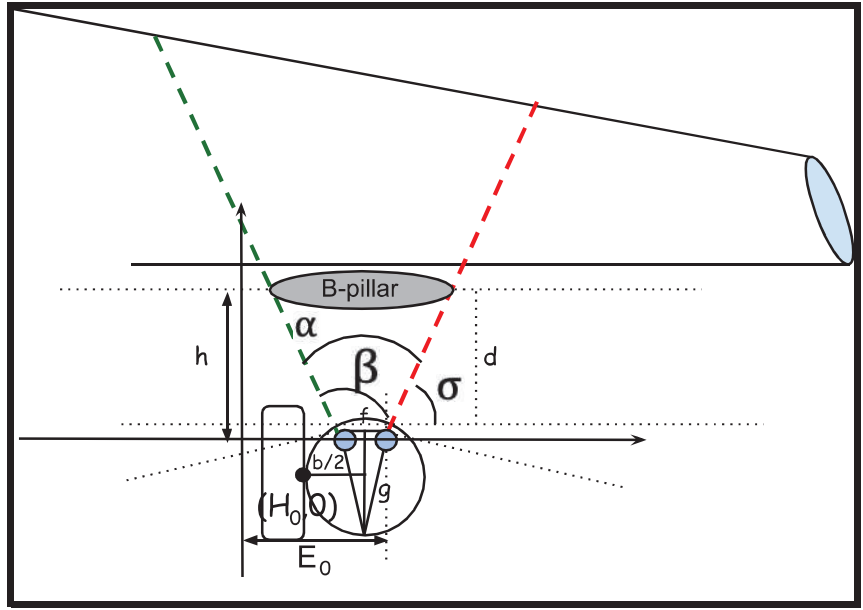

**Figure 10.** Driver checking blind spot with head turn.

To calculate the binocular obstruction angle, α, the origin is established at the intersection of the line from the left edge of the B-pillar and the line going through the driver's eyes. $\sigma$ is the angle that the right eye's obstruction line, shown in red above, makes with the B-pillar and β is the angle that the left eye's sight line makes, shown in green above, with the B-pillar. $W$ is the width of the cross-section of the pillar, $d$ is the distance of the centerline of the B-pillar from the eyes, $f$ is the distance between the eyes, and $E_0$ is the position of the right eye on the x-axis in the default location where the head is resting on the headrest and is related to $H_0$ as below. This model assumes that the human head is round, eyes do not differ in their distances to the pillar, and that the pillar is perfectly symmetrical.

$$E_0 = H_0 + \frac{b+f}{2}$$

At any given time, there are five possibilities for the relative position of the eyes concerning the B-pillar, where either the left or right eye is to the left or right of each edge of the B-pillar. These can be seen below in **Figures 11-15** with different values of $E$, position of the right eye on the *x*-axis. $W$ is the cross-sectional width of the pillar, and the blue dots are the eye locations.

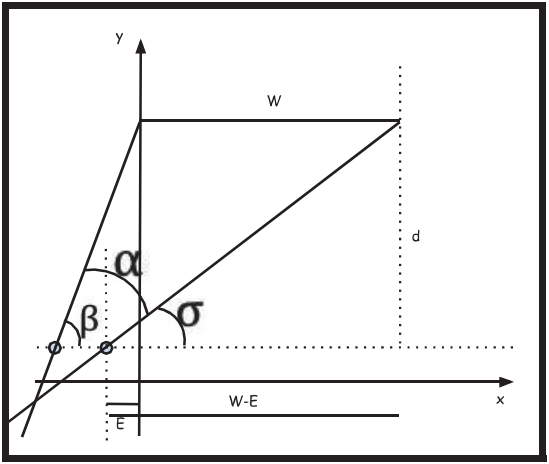

**Figure 11.** Both eyes are behind the B-pillar.

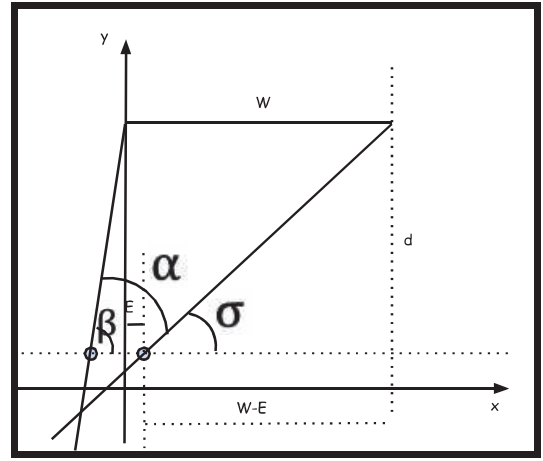

**Figure 12.** Only the left eye is behind the B-pillar.

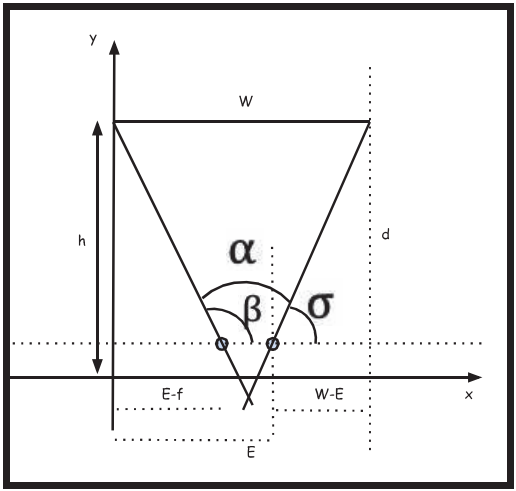

**Figure 13.** Both eyes are within the B-pillar.

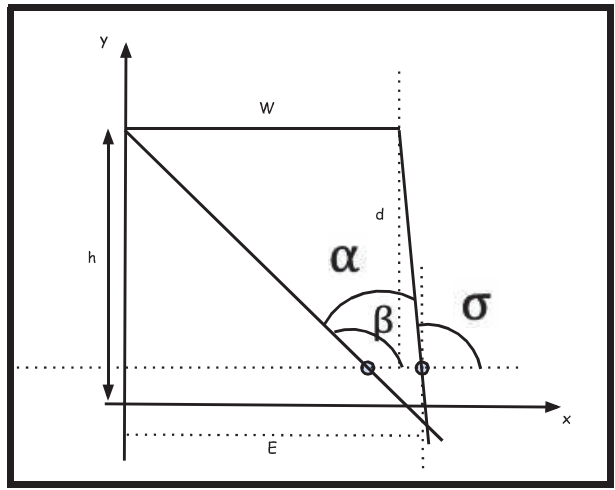

**Figure 14.** Only the right eye is ahead of the B-pillar

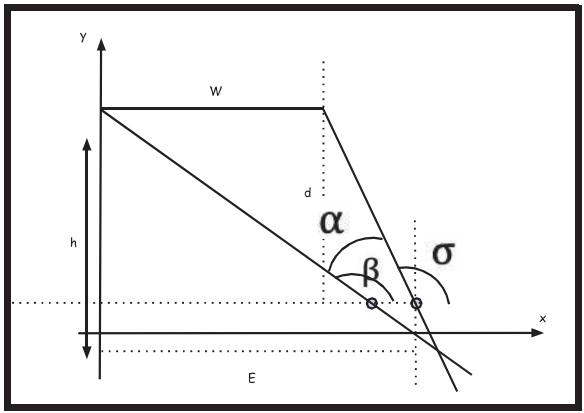

**Figure 15.** Both eyes are ahead of the B-pillar.

In each case, the obstruction angle, α, is the difference between the angles of the left and right sight lines. Thus a common equation for determining the obstruction angle for each of the above cases can be derived as follows.

$$\alpha = \beta - \sigma$$
$$\beta = arctan(\frac{d}{E-f}), \sigma = arctan(\frac{d}{W-E})$$
$$\alpha = arctan(\frac{d}{E-f}) - arctan(\frac{d}{W-E}) \ (mod \ \pi) \tag{3}$$

## 5. B-pillar blind spot with head dynamic linear motion

Subsequently, it will be determined how much head movement is necessary to ensure the blind spot from the B-pillar does not extend past the berth of what the mirror can capture, which is the sight line furthest from the car. For this derivation, this paper will ignore the sight line closest to the vehicle, as objects will only enter the area that the sight line bounds if the object first passes through the mirror view area.

This derivation assumes the driver's head is turned 90 degrees, with the driver consistently looking over their left shoulder. The first position, seen in **Figures 16** and **17** below, is the driver's head against the headrest and determines the edge of the blind angle. The second position, seen in **Figures 16** and **17** below, is the position the driver has moved to compensate for the obstruction.

In this model, the left eye's sight line is utilized for the head during the first movement, and the right eye's sight line is used during the second head movement. Successful blind spot removal is defined as the two sight lines meeting at or below the mirror sight line, as all the blind spot areas that the driver experiences will then be contained within the mirror view area. **Figure 16** below shows an example of insufficient movement, and the blind spot is reduced but not eliminated. In the next figure, **Figure 17**, the driver has moved enough to eliminate the blind spot.

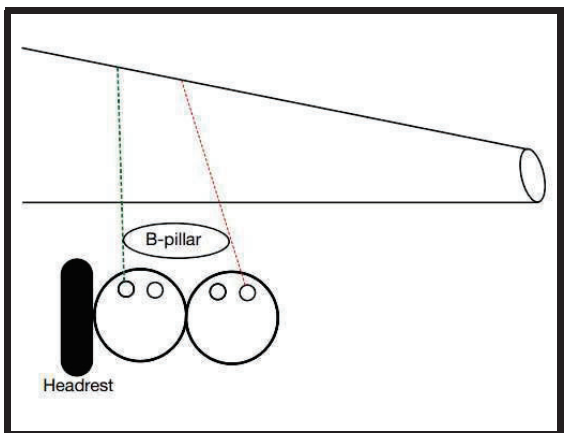
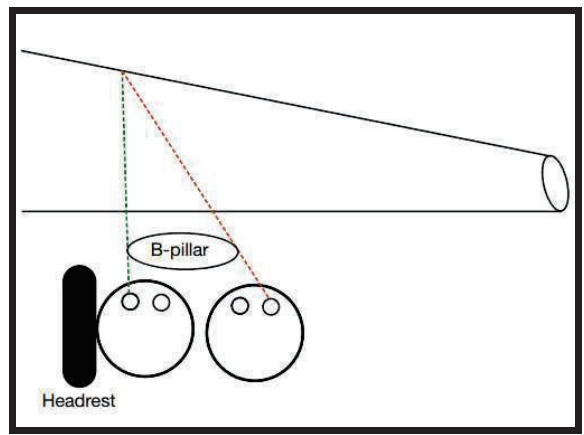

**Figure 16.** Insufficient movement and the resulting blind spot.

**Figure 17.** Sufficient movement for complete blind spot elimination.

To successfully determine the minimum distance the head needs to move to eliminate the blind spot, it is pertinent to discern at what position the mirror sight line and first position sight lines will cross. Then, the position at which the second movement's sight line must meet this intersection is found to determine the adjustment distance. To achieve this, the slope of the left eye at first and the right eye at second positions will be established.

The slope equation for the original head location can be derived from the distance of the left eye to the left corner of the pillar, utilizing the following variables outlined in section 2: $d$, the vertical distance to the pillar, $E$, the distance of the right eye to the left corner of the pillar, and $f$, the distance between the eyes. A slope equation is then developed using the triangle constructed from these bounds. This slope equation is equated to the mirror slope equation to determine the $x$ and $y$ coordinates where the two lines cross.

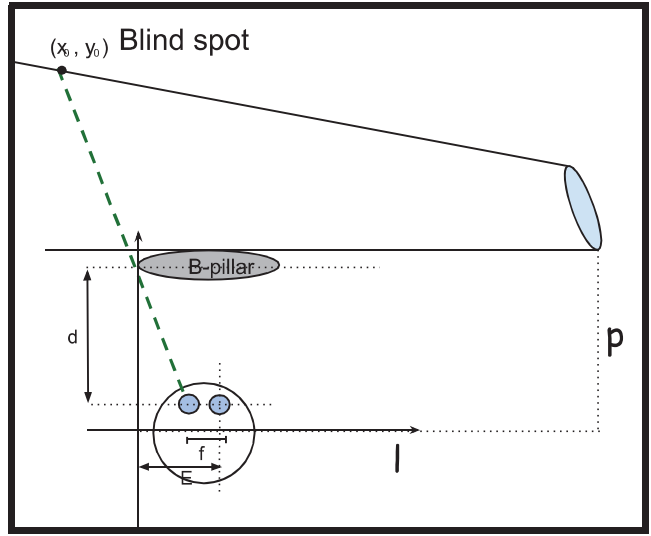

**Figure 18.** Initial position for blind spot check.

$$y_0 = \frac{d \cdot x_0}{f - E_0} + h = y_m = tan(\kappa) \cdot (l - m \cdot sin(\theta) - x_m) + m \cdot cos(\theta) + p$$

$$x_0 = x_m$$

$$x_0 = \frac{(l - m \cdot sin(\theta)) \cdot cot(2\theta + \tau_0) + m \cdot cos(\theta) + p - h}{\frac{d}{f - E_0} + cot(2\theta + \tau_0)} \tag{4}$$

Following this, the second location of the right eye is calculated by first determining the slope of the sight line, this time of the right eye. Using the slope, the sight line can be denoted with the equation below.

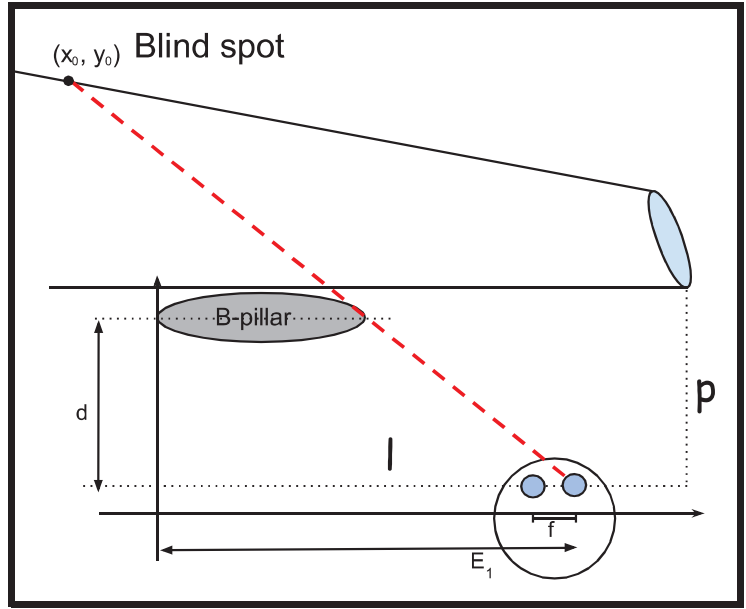

**Figure 19.** Final position for blind spot check.

$$y_1 = \frac{d}{W-E_1}(x_1 - W) + h$$

The variables $x$ and $y$, determined from the two other equations, are plugged into the slope equation for the second head movement. Once given this information, the $E$ value for the second head movement $E_1$ can be determined. The difference between $E_1$ and $E_0$, the $E$ value for the initial positions, is equal to the distance the driver must move their head to eliminate the entirety of the blind spot by ensuring the two positions' sight lines meet at the mirror sight line.

$$x_1 = x_0 = \frac{l \cdot \cot(2\theta + \tau_0) + m \cdot \cos(\theta) + p - h}{\frac{d}{f - E_0} + \cot(2 \cdot \theta + \tau_0)}$$

$$y_1 = \frac{d}{W-E_1}(x_1 - W) + h = y_0 = \frac{d \cdot x_0}{f - E_0} + h$$

$$E_1 = W + \frac{(x_0 - W) \cdot (E_0 - f)}{x_0} \tag{4a}$$

$$\Delta E = E_1 - E_0$$

$$\Delta E = W + \frac{(x_0 - W) \cdot (E_0 - f)}{x_0} - E_0 \tag{4b}$$

## 6. The Blind Spot Eliminator

Using equations (2) and (3) for a series of practical $E$ distances, the following graph is obtained using characteristics for a typical stock 2007 Honda Odyssey [32] with $d = 25.4\ cm$. Distance between the pupils for an average driver used as $f = 6.7\ cm$. In the graph, three possible head locations are considered to evaluate all the five possible relative positions for eyes as mentioned above.

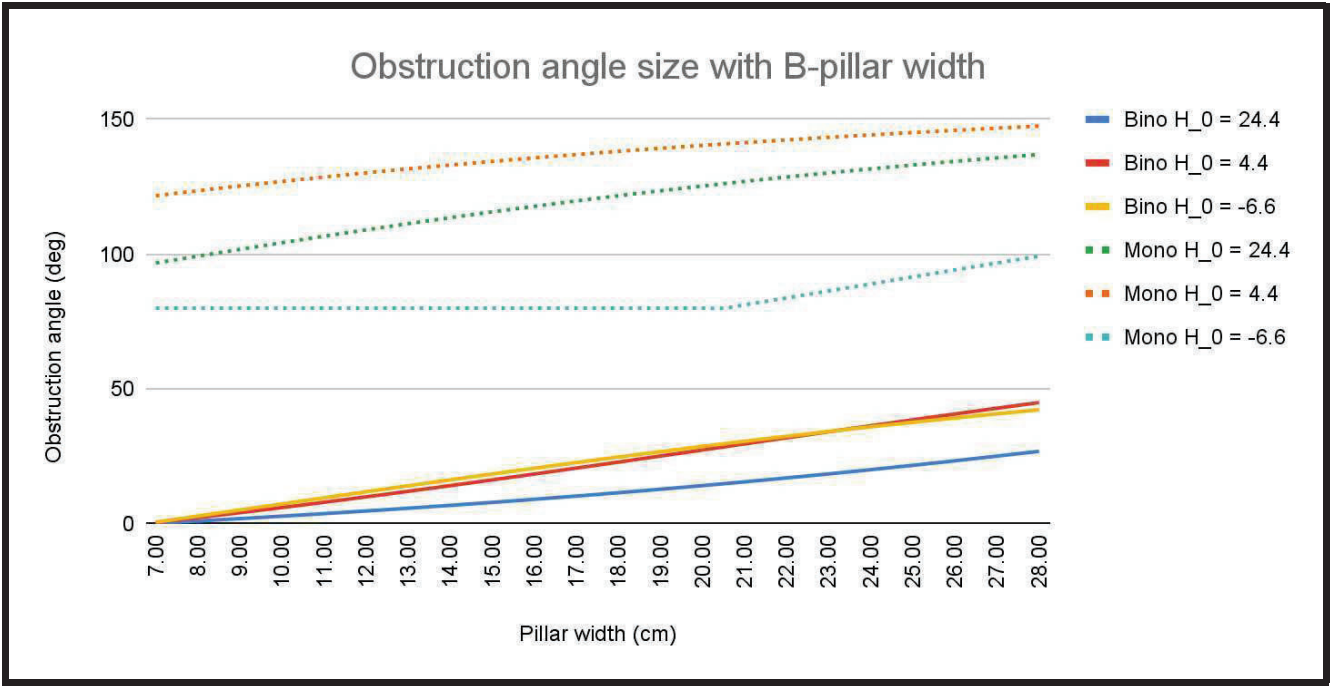

**Figure 20.** Plot showing obstruction angle with varying B-pillar width at three eye locations (cm) for monocular and binocular scenarios.

    The plot, using Equations (2) and (3) derived above, for monocular and binocular obstruction angle is shown in **Figure 20** above. The parameters used were taken from a 2007 Honda Odyssey [32] passenger van. The more diminutive the width of the B-pillar is, the smaller the corresponding obstruction angle becomes, a trend observed with any original head location. The B-pillar would need to be less than 7 cm wide to diminish the obstruction angle so that the driver experiences no impact from the pillar. This effort would be highly costly for vehicle manufacturers, as the design would almost certainly require carbon fiber or other advanced materials. As a solution, this paper proposes creating a refractive area next to the B-pillar on the driver's side window as a practical alternative. This is prototyped in this study with high-quality refractive sheet elements and is attached to the inside of the driver's side window. This view enhancer will create a refractive angle allowing drivers to see objects in the blind spots created by the driver's side B-pillar without needing to turn their heads to extreme angles or move their heads excessively. In the rest of the paper, this refractive sheet will be referred to as The Blind Spot Eliminator, or BSE.

## 7. B-pillar blind spot through BSE in default driving position

    In the following section, this paper will examine the effects of BSE on the obstruction angle experienced as the driver's head remains stationary in the default driving position. As in section 2.1, this is considered monocular vision since the area of concern is entirely out of sight for the right eye.

µ is defined as the angle of refraction for BSE and λ is the monocular obstruction angle as before, except in this case, it is the angle the refracted sight line makes with the horizontal axis, as illustrated in **Figure 21** below.

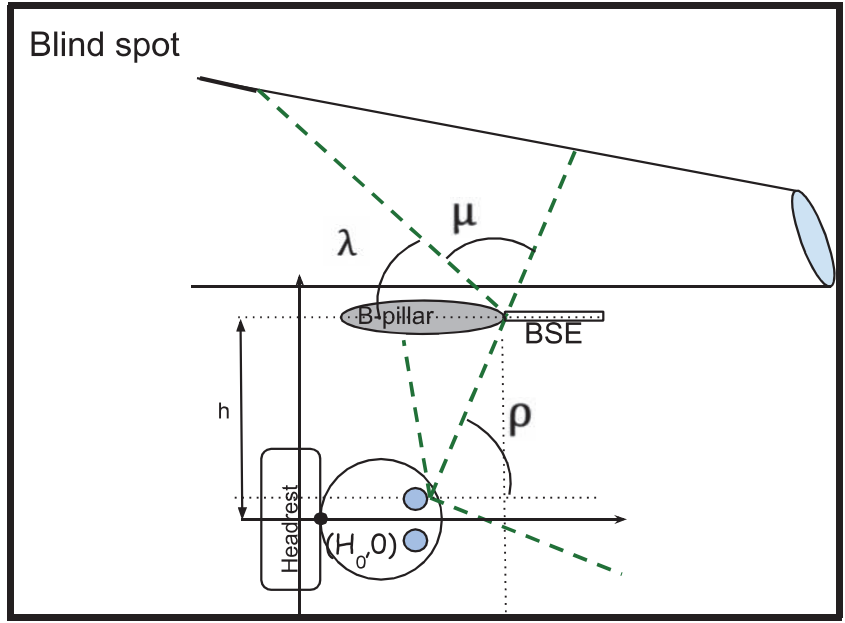

**Figure 21.** Reduced obstruction angle with BSE.

$$\rho = arctan(\frac{h-\frac{f}{2}}{W-H_0-b})$$

$$\lambda = 180 - (\rho + \mu)$$

$$\lambda = 180 - arctan(\frac{h-\frac{f}{2}}{W-H_0-b}) - \mu = arctan(\frac{h-\frac{f}{2}}{H_0+b-W}) - \mu \ (mod \ \pi) \quad (5)$$

With the equations (1) and (5), it will be possible to consider different positions for the driver's head and assess the effective obstruction experienced by the driver, with and without the help of BSE. Since human peripheral vision extends to 10 degrees on the left, as shown in **Figure 1**, the smallest value possible for the unassisted obstruction angle is 80 degrees. With BSE, it is possible to reduce the angle further and reduce the blind area.

The differences between the magnitude of the obstruction angle when the driver utilizes solely monocular vision between the calculation done without BSE, section 2.1, and with BSE, section 4, is quite stark. In this scenario, the refractive angle of the BSE addition was 45.7 degrees. When examining the following graphs with distance variables adjusted, a clear trend emerges of the subsequent obstruction angle.

Using equations (1) and (5) with a series of driver's head positions, the following graph is obtained using characteristics for a stock 2007 Honda Odyssey [32] with h = 25.40 cm, f = 6.70 cm, $H_0$ = 20cm, g = 19, and b = 18.70.

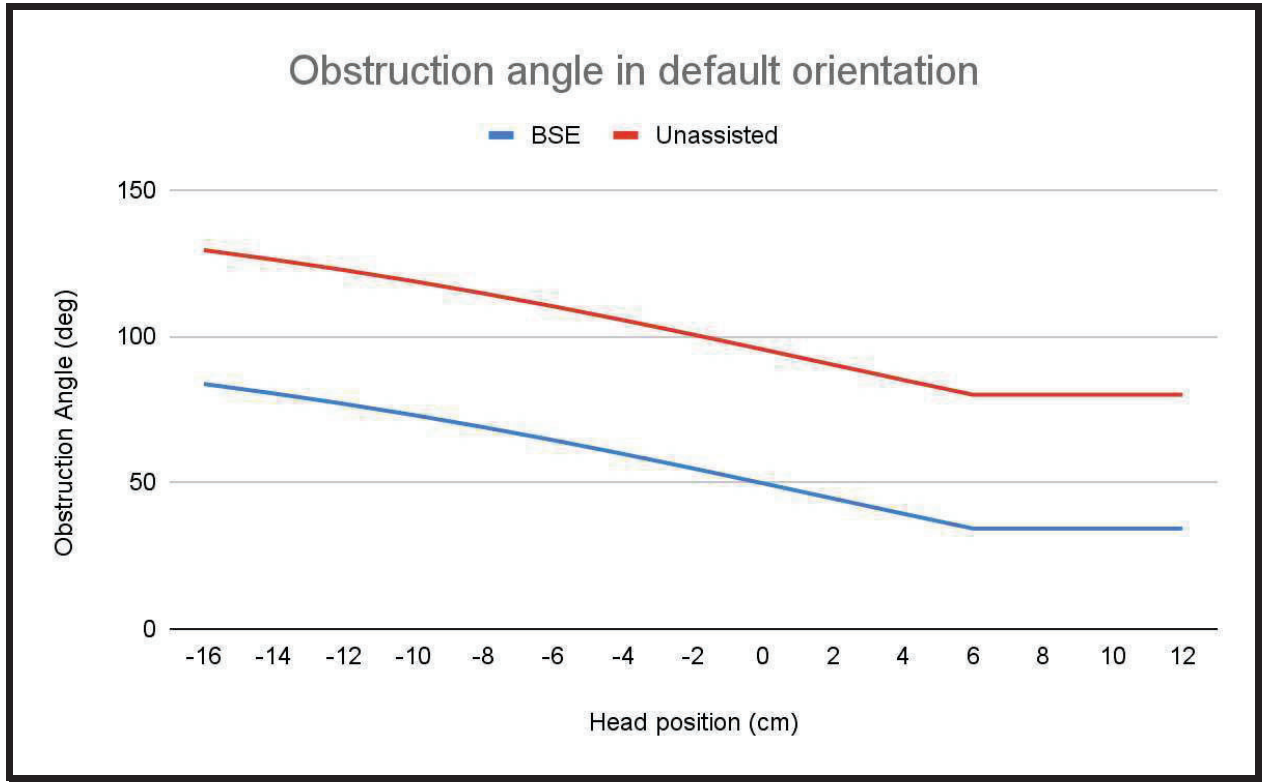

**Figure 22.** Plot for obstruction angle with different head positions.

As seen in the graph above, the contrast in the magnitude of the obstruction angles in the cases with and without BSE demonstrates that the case with BSE had an obstruction angle that was reduced by µ degrees, a reduction of, on average, 41.37 %. In addition, the obstruction angle with the utilization of BSE will reach a minimum of 34 degrees when $H = 6$ cm. In comparison, the angle of obstruction without BSE will remain at 80 degrees, the edge of monocular vision, as shown in **Figure 21**. Thus, the evidence shows that the BSE dramatically reduces the obstruction angle of the driver in all cases, no matter the original position.

## 8. B-pillar blind spot through BSE with head turn

Here, this study will derive an equation determining the distance needed to travel with the addition of BSE in the following section. The next set of calculations uses the starting point for the movement as the location of the right eye at the furthest back against the headrest. Successful blind spot elimination is when two sight lines of the individual head positions meet at or before the sight line of the mirror, as was done previously, this time with the refraction angle, µ, illustrated below in **Figure 23**.

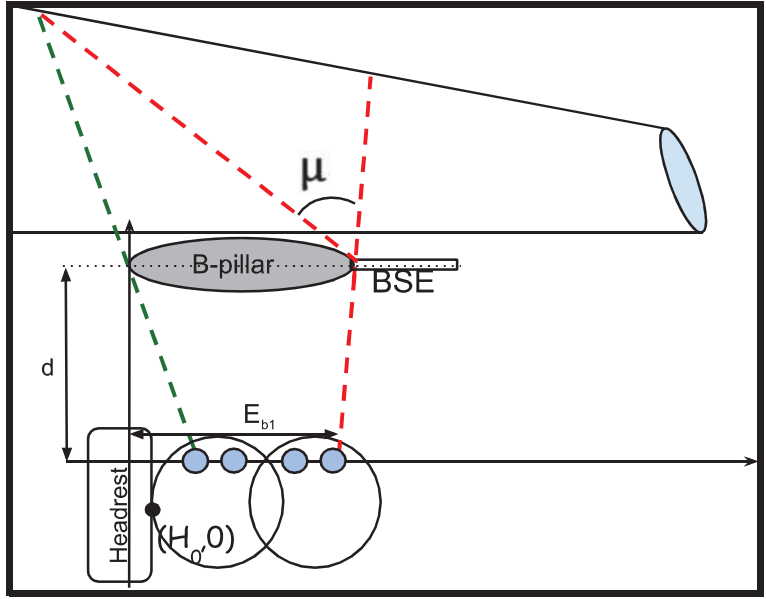

**Figure 23.** Head movement to eliminate blind spot with BSE.

Compared to section 3, which also calculated the distance the head needed to travel to remove the blind spot entirely, the only difference in this section is that a refractive sheet now assists the secondary, or adjustment, head movement with a refractive angle μ. Thus, the process and equations for determining the mirror sight line and the first head location's sight line crossing point can remain the same.

The head adjustment derivation is the same for both cases. By writing a slope equation of the final angle of the eye line with angle λ, affected by the refractive angle of the sheet, the location the head needs to be at to meet the other two sight lines can be determined.

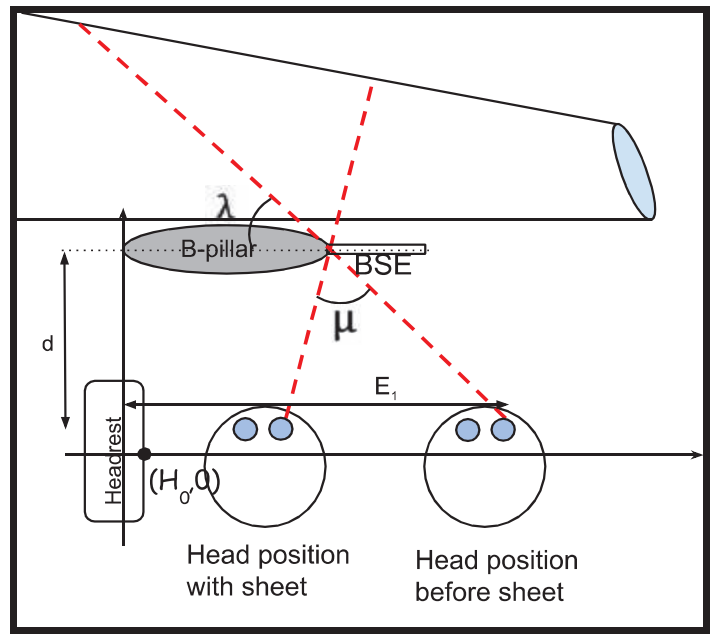

**Figure 24.** Difference in head movement with BSE.

As the equation for the refracted line has now been derived to determine the required head relocation to cover the same blind spot, the same process as in section 3 can be utilized.

Since the driver's head is initially on the headrest at its farthest position back, the intersection of the sight line for the left eye and the outer mirror line is still our starting point. Coordinates for this point were referred to as $(x_0, y_0)$ previously. $x_0$ was derived from the original right eye location $E_0$ as below.

$$x_0 = \frac{l \cdot tan(\kappa) + m \cdot cos(\theta) + p - h}{\frac{d}{f - E_0} + tan(\kappa)}$$

From there, the values for the $x_0$ and $y_0$ locations determined in the previous step will be plugged into the equation for the slope of the right eyesight line.

$$y_1 = x_1 \cdot tan(arctan(\frac{d}{W - E_{b1}}) + \mu) + h$$

$$y_1 = (x_1 - W) \cdot tan(arctan(\frac{d}{W - E_{b1}}) + \mu) + h = y_0 = \frac{d \cdot x_0}{f - E_0} + h$$

$$tan(arctan(\frac{d}{W - E_{b1}}) + \mu) = \frac{d \cdot x_0}{(x_0 - W) \cdot (f - E_0)}$$

$$if\ v = \frac{d}{W - E_{b1}}, w = \frac{d}{(f - E_0)}, then, tan(arctan(v) + \mu) = w \cdot \frac{x_0}{x_0 - W}$$

Using the identity $tan(a + b) = \frac{tan(a) + tan(b)}{1 - tan(a)tan(b)}$

$$\frac{a + tan(\mu)}{tan(\mu) - a} = b \cdot \frac{x_0}{x_0 - W} \Rightarrow a = \frac{b \cdot \frac{x_0}{x_0 - W} - tan(\mu)}{tan(\mu) + b \cdot \frac{x_0}{x_0 - W}}$$

$$\frac{d}{W - E_{b1}} = \frac{b \cdot \frac{x_0}{x_0 - W} - tan(\mu)}{tan(\mu) + b \cdot \frac{x_0}{x_0 - W}}$$

$$E_{b1} = W - \frac{d + \frac{tan(\mu) \cdot d^2 \cdot x_0}{(x_0 - W) \cdot (f - E_0)}}{\frac{d \cdot x_0}{(x_0 - W) \cdot (f - E_0)} - tan(\mu)} \tag{6a}$$

The distance moved by the head, $\Delta E_b$, is defined as $E_{b1} - E_0$. $\Delta E_b$ for the above case is shown below.

$$\Delta E_b = E_{b1} - E_0 = W - \frac{d + \frac{tan(\mu) \cdot d^2 \cdot x_0}{(x_0 - W) \cdot (f - E_0)}}{\frac{d \cdot x_0}{(x_0 - W) \cdot (f - E_0)} - tan(\mu)} - E_0 \tag{6b}$$

Next, the difference between the distance required for an individual to move their head to eliminate the blind spot with and without BSE will be examined. The following graphs can be discerned from cases where BSE is not used, where the angle of refraction for BSE is 30 degrees, and where the angle of refraction for BSE is 45.7 degrees.

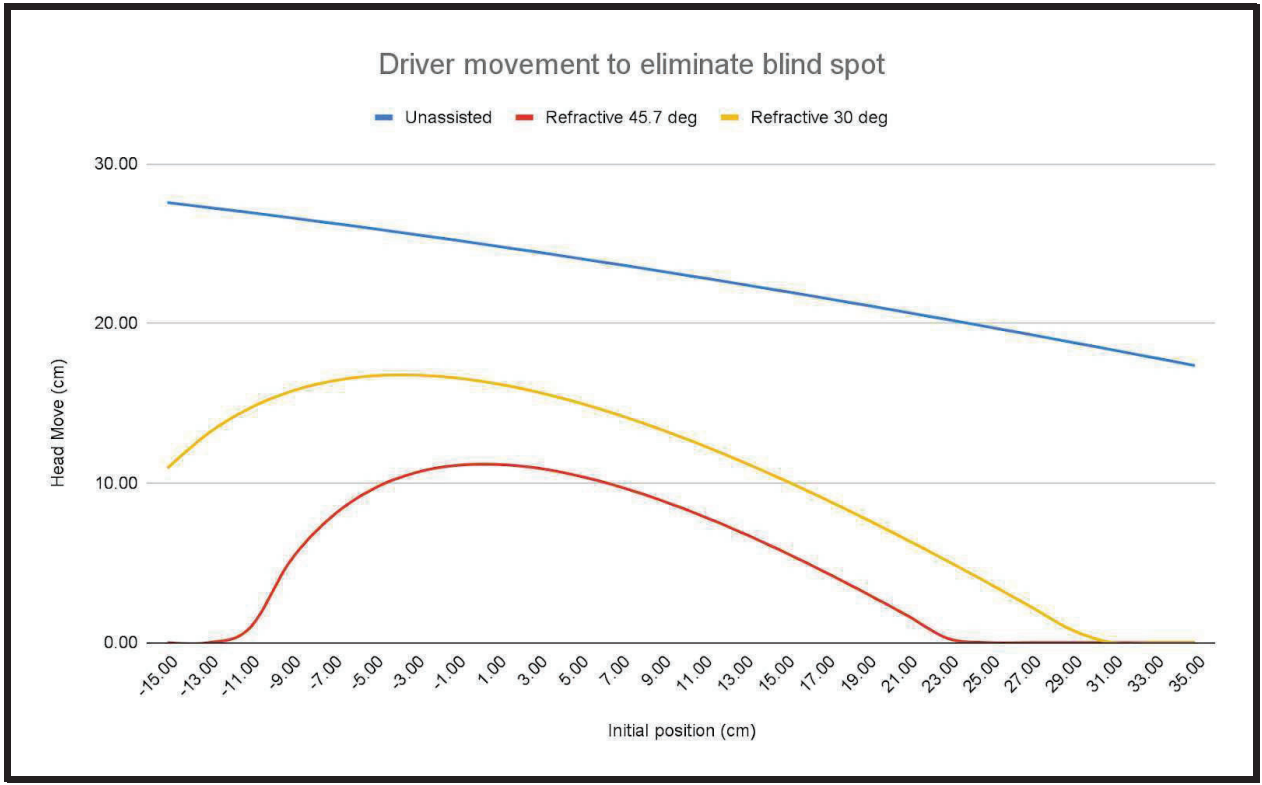

**Figure 25.** Head movement required to eliminate blind spot.

As observed, the distance required to move the head to eliminate the blind spot without BSE is more extensive than the distance needed when BSE has a refractive property of 30 degrees and is summarily more extensive when the refractive property is 45.7 degrees. With a refractive property of 30 degrees, the distance needed to remove the blind spot was reduced by an average of 58.11%, and the refractive property of 45.7 degrees led to a reduction of 79.85% on average. Thus, these values provide a persuasive argument for the utilization of BSE, due to the extreme reduction of the distance required to move when BSE is utilized.

## 9. The Blind Spot Eliminator Prototype

To demonstrate the effect of BSE as described above, with a real-world situation, a comparison will be made between the view of a model with and without a refractive sheet, such is modeled in sections 4 to 6.

The vehicle used for this experiment was a 2007 Honda Odyssey [32]. This vehicle has a considerably large B-pillar, measured at $W = 20.8\ cm$, as the seat belt is located within the pillar. The distance from the pillar to the middle of the headrest is $h = 25.4\ cm$. The driver's side mirror width is $m = 17.8\ cm$, it is angled with $\theta = 9°$ from perpendicular to the side of the vehicle. The base of the mirror is slightly farther from the headrest with $p = 27.9\ cm$ with the distance from the left edge of the B-pillar $l = 107\ cm$.

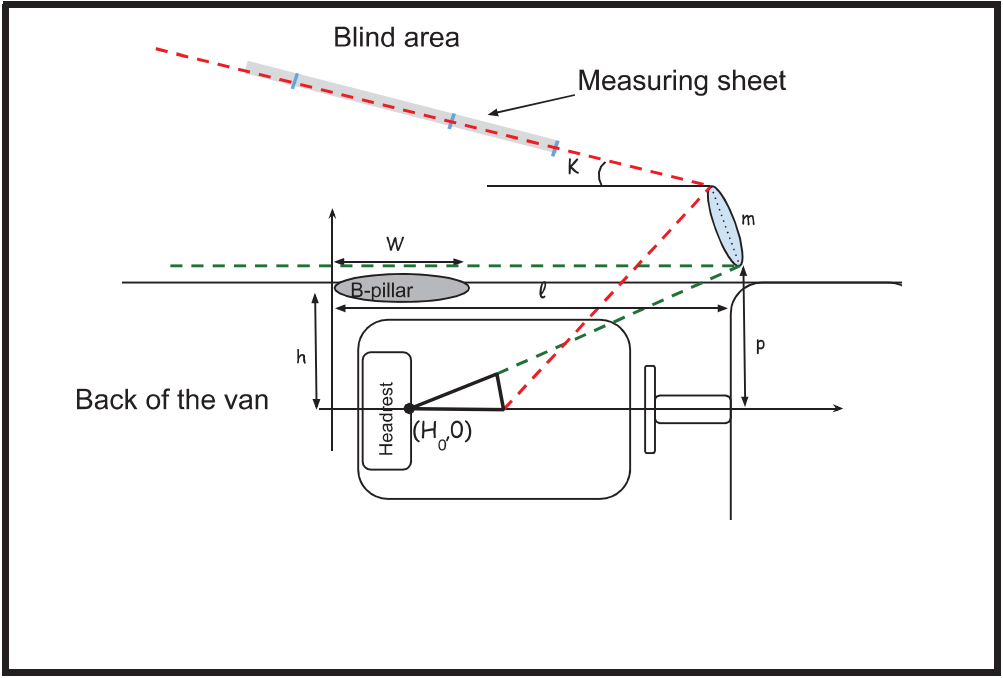

**Figure 26.** Layout of the experimental setup.

    For the following experiments, a mannequin was used with the appropriate eye distance characteristics. The sight lines were simulated with lasers, green for the left and red for the right eyes. They were attached to a rotating metal wand, and the pivots of these wands representing pupils of the model were located $f = 6.7 cm$ from each other, an interpupillary distance within the statistical range for adults [29]. Pupils for the model were $g = 19 cm$ from the center point on the back of the head.

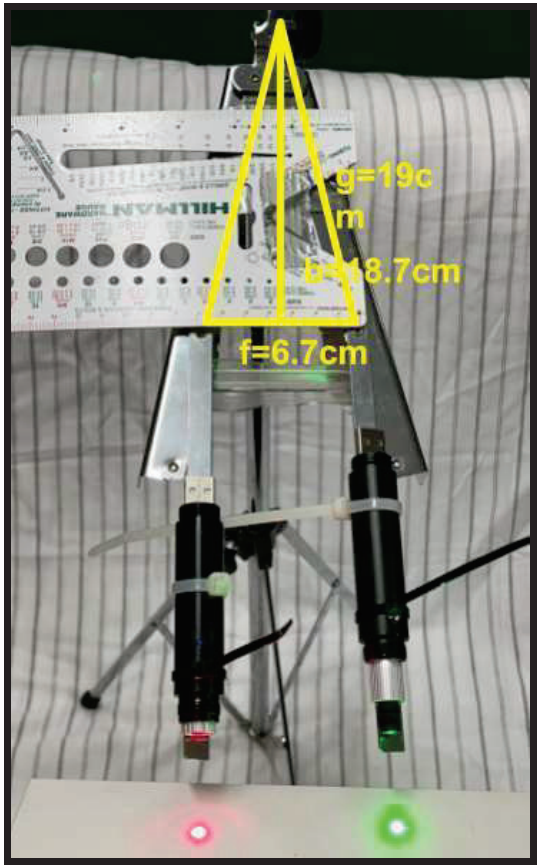

**Figure 27.** Mannequin set up to look straight ahead.

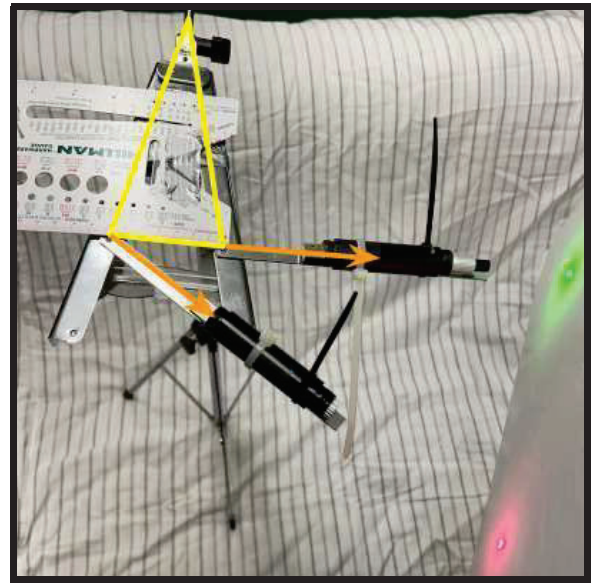

**Figure 28.** Mannequin looking left with the right eye at its maximum turn.

    The model's head was placed in the seat at a height of 71 *cm*, in the typical driver's default position of the head resting at the headrest at a position $H_0 = 6$ *cm*.

    As a first step, the model's eyes were angled to view the driver's side mirror, as shown in Image 3 and 4 below. Utilizing the lasers to represent the extent of the driver's vision through the side mirror, a large measuring sheet was hung and angled in the exact location as the model's furthest sight line to show the bounds of the area a side mirror would cover for a typical driver. The area behind the sheet, not in the periphery of the driver's vision, is considered the blind area.

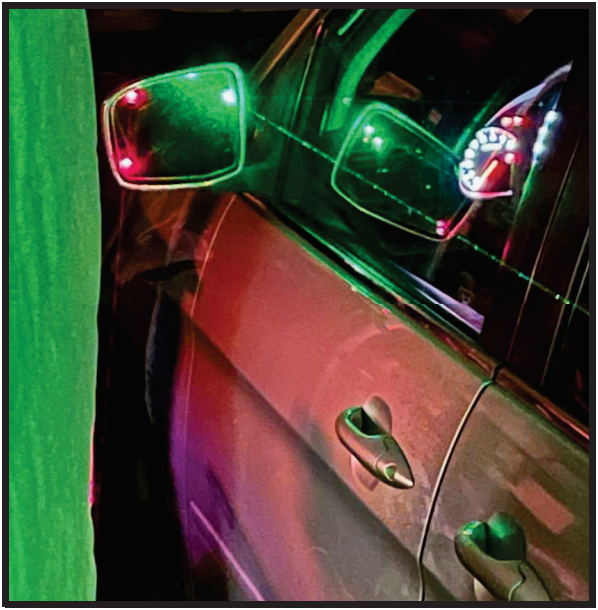 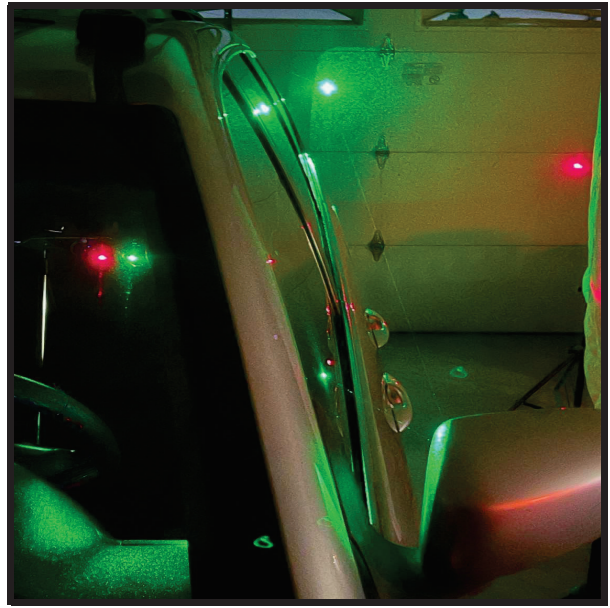

**Figure 29.** Laser lines reflected by the side mirror.

**Figure 30.** View area covered by mirror.

With the parameters listed above, the mirror sight line was established as $\kappa = 10.3°$.

### 9.1. - Monocular vision in driver's default position

This section compares the monocular viewing of the blind area with and without BSE. This section can be considered the real-life depiction of section 4. An improved result for this experiment is defined as the sight line moving further down the sheet, thus reducing more of the obstruction angle.

The model was leaned back with the focal point of its eyes located at the headrest, the equivalent of a driver putting their head against the headrest to drive, which is considered within the range of safe head positions [33]. The left sight line, the green laser, was angled to brush against the B-pillar, with the laser angled within the 120-degree range of the eye.

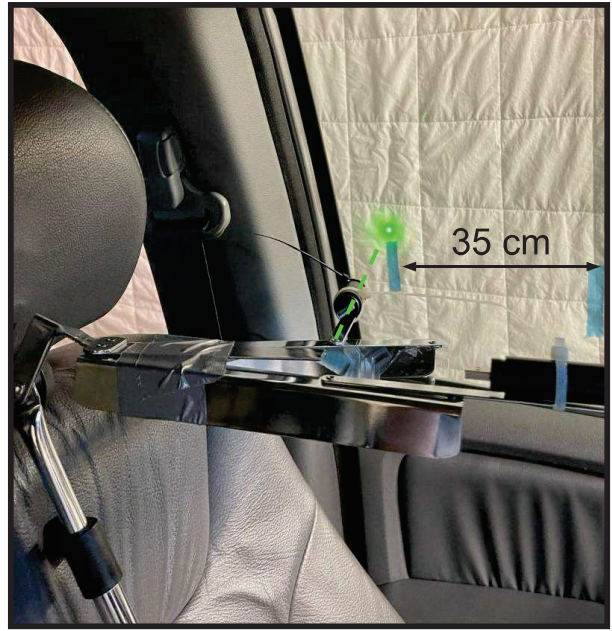

**Figure 31.** Driver's left eye with limited coverage on mirror sight line.

Since only the left eye covers the left side of the vehicle in this scenario, the area behind the laser line seen to the right side of the B-pillar is considered blind. This point was measured to be 35 cm from the reference line on the right in Image 5 above.

## 9.2. - Monocular vision with BSE

For the second portion of this experiment, BSE is attached to the window, and the model's eye is left unaltered, the laser line shining immediately to the right of the B-pillar as before.

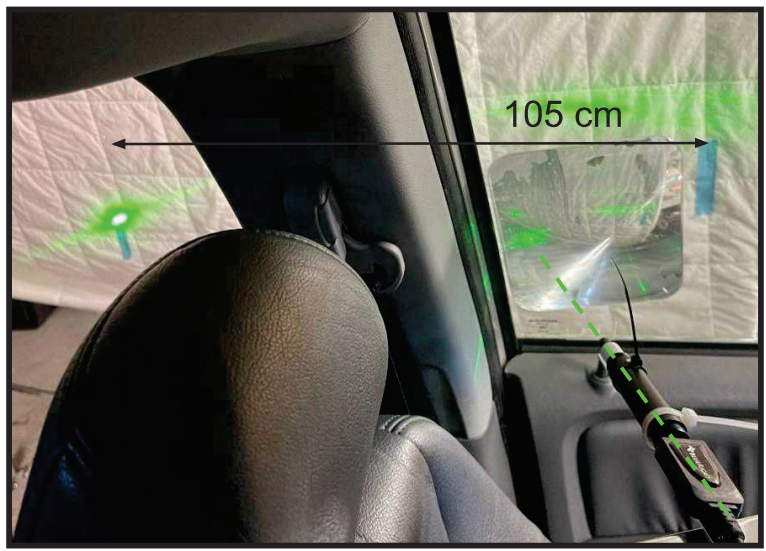

**Figure 32.** Driver's left eyesight line being aided by BSE.

As seen in Image 6 above, the laser sight line was significantly altered and is not pointing to a different point further back on the measuring sheet. The new location of the sight line was measured to be 105 cm from the reference line on the sheet.

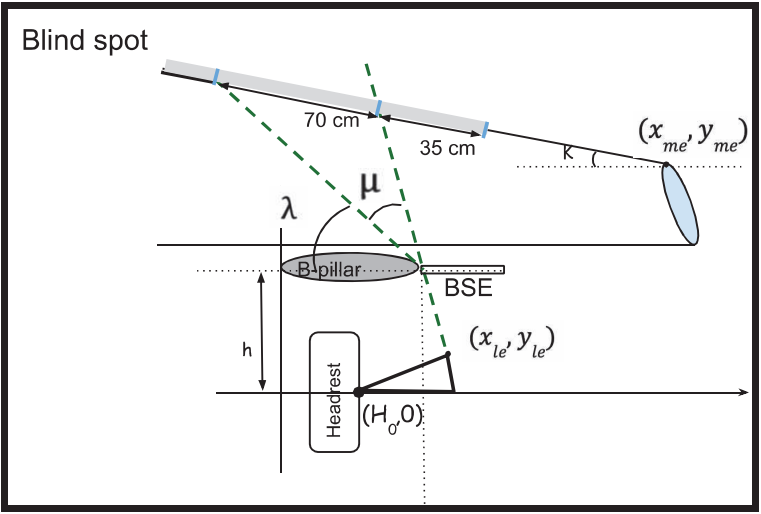

**Figure 33.** Monocular vision enhancement with BSE.

The sight lines before and after can be seen in **Figure 33** above. By using the reference points $(x_{me}, y_{me})$ and $(x_{le}, y_{le})$, we can accurately estimate the refraction angle achieved by BSE.

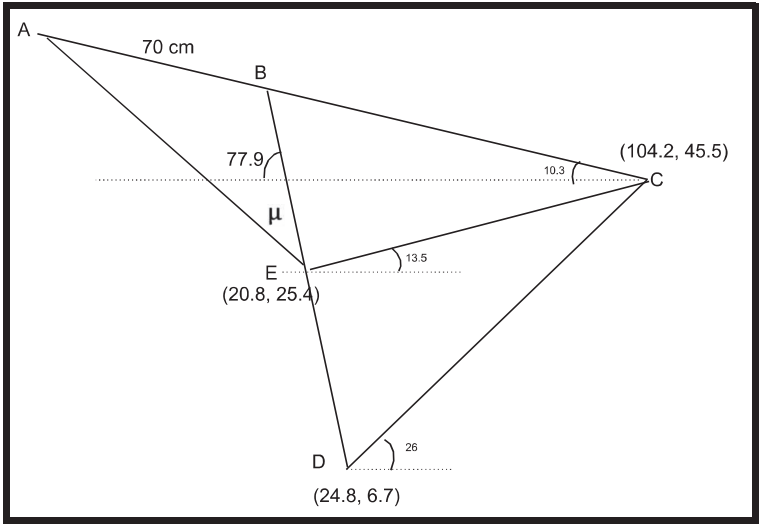

**Figure 34.** Geometrical representation of the driver's left eye and the mirror line.

The geometrical representation of the experiment as shown in **Figure 34** can be used to walk through the calculations for μ. The mirror sight line is at an angle $\kappa = 10.3^0$, starting from the edge of the mirror at position

$$(x_{me}, y_{me}) = (l - m \cdot sin(\theta), p + m \cdot cos(\theta)) = (104.2, 45.5).$$

Using $H_0$, $f$ and $g$ values mentioned earlier, it can be determined that the model's left eye is at the position $(x_{le}, y_{le}) = (23.8, 6.7)$, and the angle it makes with the horizontal axis is:

$$\lambda + \mu = arctan(\frac{25.4-6.7}{24.8-20}) = 77.9°$$

We can calculate the other angles and line lengths as:

$$\angle EBC = 77.9° - \kappa = 67.6°$$
$$|EC| = 85.8\ cm,\ \angle ECB = \kappa + 13.5° = 23.8°$$
$$thus\ by\ Law\ of\ Sines,\ |BE| = \frac{89.5 \cdot sin(23.3)}{sin(68.2)} = 37.5\ cm.$$
$$Since\ |AB|\ is\ 70\ cm\ and\ \angle EBA = 112.4°,\ \mu\ is\ 45.7°$$

As shown, the difference between the two sight lines is quite significant, as an additional 70 cm of the mirror sight line edge is in view by utilizing BSE. The obstruction angle decreases by 45.7 degrees. Therefore, in this model, adding BSE significantly reduces the blind spot and correspondingly improves the safety of the driver.

## 10. Results

As seen in **Figure 20**, the obstruction angle increases rapidly as the width of the B-pillar increases, and to achieve no negative effect B-pillar width would need to be 7 cm or less. Drivers would need to move significantly to eliminate the blind spots, as determined by Equation 4b. The proposed Blind Spot Eliminator reduces the obstruction angle and movement significantly. In the default forward-facing position, drivers using a BSE with a refractive angle of 45.7 degrees would reduce the obstruction angle for monocular vision by an average of 41.39%. This trend can be viewed in **Figure 22**. The distance the driver is required to move to eliminate a side blind spot with a BSE of 30 degrees was reduced by an average of 58.11%. In comparison, the utilization of BSE with a refraction angle of 45.7 degrees reduced the distance by 79.85%, as shown in **Figure 25**. The real-life experiment was conducted in a 2007 Honda Odyssey [32], utilizing a BSE prototype with a refractive angle of 45.7 degrees, reducing the blind spot by 70 cm and the obstruction angle by 45.7 degrees.

## 11. Conclusion

Driving has been an integral part of human lives for decades, and methods to reduce and eliminate road accidents must be explored and implemented in our vehicles. One of the major contributors to vehicular accidents is vision gaps experienced by the driver. B-pillars that support the roof of the vehicle are frequently a significant contributor to these gaps, as not only are they wide to support the weight of the car in case of a rollover, but they are also the closest pillar to the driver.

This study derived equations to quantify the obstruction angle for the two typical driving situations when the driver's head is straight along the direction of travel and when the head is turned 90 degrees to the left, as commonly done to check for objects in areas the side mirror does not cover. The study demonstrates that the obstruction due to a nominal B-pillar is quite severe. Also shown is that the driver must move forward extensively to eliminate the blind spots.

As part of this study, an assessment was created for the impact of a refractive sheet attached to the driver's side window to help the driver view areas directly behind the B-pillar. The equations derived in the study were adapted to incorporate the refractive sheet for the obstruction angle and the range of motion needed for the driver to eliminate the blind spot with a head turn of 90 degrees.

Through data analysis with varying characteristics like B-pillar size, driver's default seat, and head position, the study demonstrates the improved coverage of blind spots using the refractive sheet. The refractive sheet significantly reduced obstruction angles in the default and turned positions and reduced or eliminated the forward motion required.

In addition to the computational data collected in the study, a prototype refractive sheet, the Blind Spot Eliminator (BSE), was constructed from commercially available subcomponents. The study demonstrated improved blind spot coverage using a manikin modeling human binocular vision. The results from the real-world experiment using BSE with a 45.7-degree refraction angle were provided as a proof of concept for significantly reducing blind spots due to the B-pillar.

Future research to gather data and utilize the equations derived in this paper on the most popular vehicles in America will prove valuable to provide an informative empirical metric about the characteristics of the vehicles' blind spots. This research will improve the purchasing experience for consumers and assist vehicle manufacturers in empirically quantifying the blind spots of their vehicles, thus improving overall road safety in America.